\documentclass[letterpaper, 10 pt, conference]{ieeeconf}  

\IEEEoverridecommandlockouts                              

\usepackage{amsmath} 
\usepackage{amssymb}  

\usepackage{graphicx}
\usepackage{tabu}
\usepackage{diagbox}
\usepackage{xcolor}
\usepackage{soul}
\usepackage{caption}
\title{\LARGE \bf
Read My Mind: A Multi-Modal Dataset for Human Belief Prediction. 
}

\author{Jiafei Duan$^{1}$, Samson Yu$^{2}$, Nicholas Tan$^{2}$, Yi Ru Wang$^{1}$ and Cheston Tan$^{3}$%
\thanks{$^{1}$University of Washington, USA, {\tt\small duanj1@cs.washington.edu}}%
\thanks{$^{2}$National University of Singapore, Singapore}%
\thanks{$^{3}$Centre for Frontier AI Research, A*STAR, Singapore}%
}

\begin{document}

\maketitle
\thispagestyle{empty}
\pagestyle{empty}

\begin{abstract}
Understanding human intentions is key to enabling effective and efficient human-robot interaction (HRI) in collaborative settings. To enable developments and evaluation of the ability of artificial intelligence (AI) systems to infer human beliefs, we introduce a large-scale multi-modal video dataset for intent prediction based on object-context relations. 

\end{abstract}

    


\section{Introduction}
Humans with an average level of social cognition can infer the beliefs of others solely based on the nonverbal communication signals (e.g., gaze, gesture, pose, and contextual information) displayed during social interactions \cite{hinde1974biological,wellman2001meta,saxe2006uniquely,wang2023study}. Recent technological advancements in robotics and AI have increased the demand for human-robot interaction (HRI) \cite{duan2022survey} in various domains (e.g., manufacturing, service, healthcare, etc.). Central to understanding human intents is perception of human actions during interactions, and an understanding of object-context relations, defined as the knowledge about objects occurring in a given context. Hence, we propose a novel video dataset with multi-modal annotations which captures human behaviours in an object-context setting.


\section{Data Collection}
\begin{figure}[h]
    \centering
    \includegraphics[width=\linewidth]{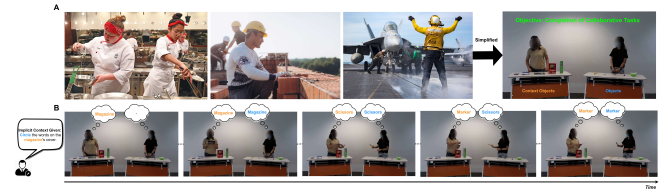}
    \caption{(A) Real-world examples of collaborative tasks that require belief inference via nonverbal communication. (B) Example of instructions provided during data collection.}
\label{fig:0}
\end{figure}

Our dataset includes 10 pairs of participants—five pairs of friends and five pairs of strangers—from 15 different contexts. We collected 347,490 frames from 900 egocentric and third-person videos. As shown in Figure \ref{fig:0}B, each individual within participant pairs are presented with either contextual objects (demonstrator) or context-specific tools (predictor). The objects and tools are commonly found household objects, and most can be found in the YCB dataset \cite{calli2015ycb}. The demonstrator will be given an implicit context task, and will choose a contextual object based on the task. The individual will then convey the implicit context task non-verbally using the chosen object to the predictor, who will then choose a tool from the list of context-specific tools by inferring the demonstrator's intent. 


Throughout the experimental process, participants wore noise-cancelling headphones and verbalized the names of items they have in mind whenever their beliefs are updated. This enabled accurate annotation of the participants' hidden beliefs in relation to the frame of the captured videos. We also gathered hand gesture data from two Leap Motion sensors. Other input modalities, including object detection, pose estimation, and gaze tracking, were accomplished through a post-processing approach. We use Detecto \cite{Detecto} to detect objects with further fine-tuning on our self-annotated frames, Gaze360 \cite{kellnhofer2019gaze360} to collect 3D human gazes, and OpenPose \cite{cao2017realtime} to obtain all the critical points for posture estimation. 







\begin{figure}
    \centering
    \includegraphics[width=0.8\linewidth]{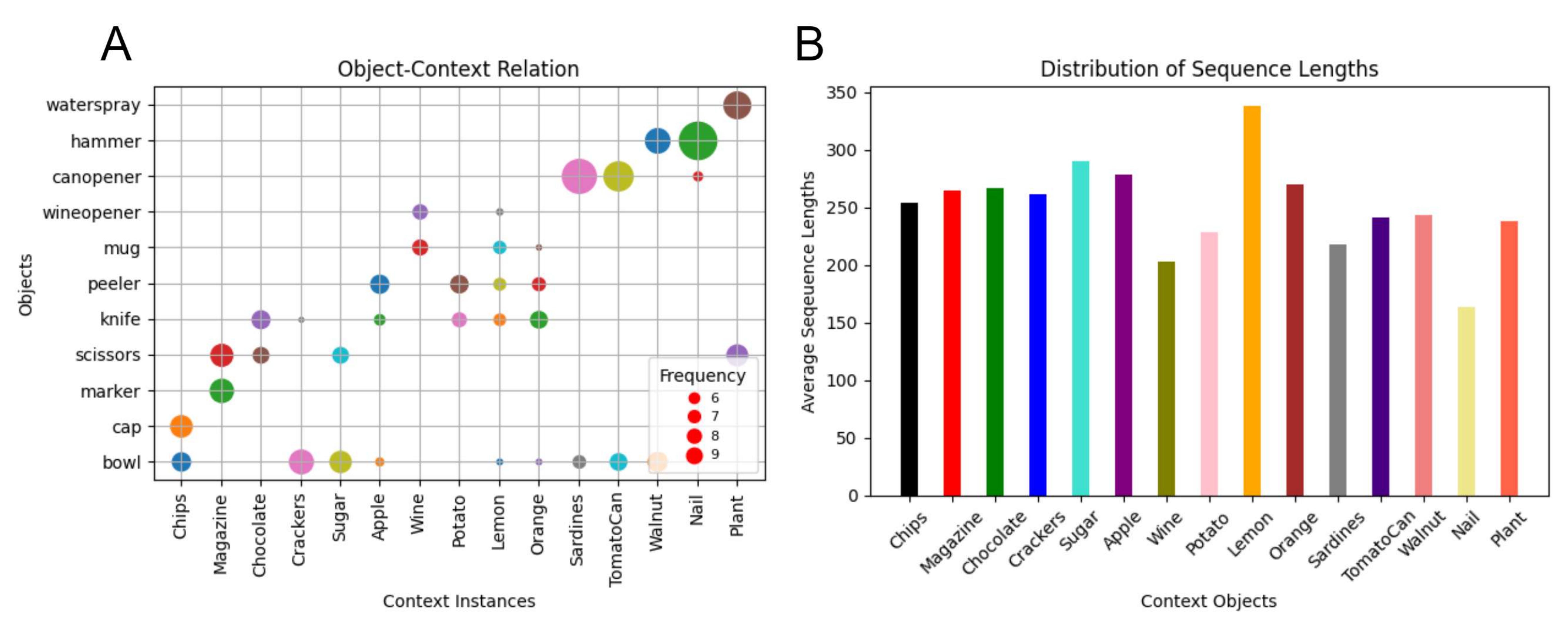}
    
    \caption{Overview of the dataset statistics. (A) The frequency of all the potential object-context matches. (B) The distribution of sequence lengths across the context object classes. }
\label{fig4}
\end{figure}

\section{Dataset analysis}
The dataset contains 900 recorded videos captured concurrently from both egocentric third-person views, yielding a total of 347,490 annotated frames or approximately 3.2 hours of video. As depicted in Figure \ref{fig4}B, the average duration of a context object sequence is around 8 seconds, or 250 frames at a frame rate of 30fps. All videos were acquired with audio that was synced. In addition, Figure \ref{fig4}A demonstrates the matching frequency and probable object-context pairs for all participant pairs. Using the data received from the training set, we can create an Object-Context Relation (OCR) matrix that precisely maps out the interconnected context and object relationships. This matrix can also be utilized in training as a type of previous knowledge of the context and object's functionality or utility relationship.

\section{Conclusion}
We introduced a challenging multi-modal dataset to enable human belief prediction from videos in object-context settings. We hope that this work can facilitate future research in understanding human beliefs and contribute to the field of human-robot interaction.


\bibliography{references}

\end{document}